\documentclass{ieeeaccess}
\usepackage{cite}
\usepackage{amsmath,amssymb,amsfonts}
\usepackage{algorithm}
\usepackage[noend]{algpseudocode}

\DeclareMathOperator*{\Max}{max}
\DeclareMathOperator*{\Min}{min}
\DeclareMathOperator*{\minimize}{minimize}
\usepackage[caption=false,font=normalsize,labelfont=sf,textfont=sf]{subfig}
\usepackage{graphicx}
\usepackage{stfloats}
\usepackage{textcomp}
\usepackage{gensymb}
\usepackage{booktabs}
\usepackage{multirow}
\usepackage{bm}
\usepackage{siunitx}
\def\BibTeX{{\rm B\kern-.05em{\sc i\kern-.025em b}\kern-.08em
    T\kern-.1667em\lower.7ex\hbox{E}\kern-.125emX}}
\begin{document}
\history{Date of publication xxxx 00, 0000, date of current version xxxx 00, 0000.}
\doi{10.1109/ACCESS.2023.0322000}

\title{Adversary Aware Continual Learning}
\author{\uppercase{Muhammad Umer}\authorrefmark{1}, \IEEEmembership{Member, IEEE},
\uppercase{Robi Polikar}\authorrefmark{2},
\IEEEmembership{Member, IEEE}}

\address[1]{Department of Electrical \& Computer Engineering, Glassboro, NJ 08028 USA (e-mail: umerm5@rowan.edu)}
\address[2]{Department of Electrical \& Computer Engineering, Glassboro, NJ 08028 USA (e-mail: polikar@rowan.edu)}

\markboth
{Umer \headeretal: Preparation of Papers for IEEE TRANSACTIONS and JOURNALS}
{Umer \headeretal: Preparation of Papers for IEEE TRANSACTIONS and JOURNALS}

\corresp{Corresponding author: Robi Polikar (e-mail: polikar@rowan.edu).}

\begin{abstract}
Continual learning approaches are useful as they help the model to learn new information (classes) sequentially, while also retaining the previously acquired information (classes). However, it has been shown that such approaches are extremely vulnerable to the adversarial backdoor attacks, where an intelligent adversary can introduce small amount of \textit{misinformation} to the model in the form of \textit{imperceptible backdoor pattern} during training to cause deliberate forgetting of a specific task or class at test time. In this work, we propose a novel defensive framework to counter such an insidious attack where, we use the attacker’s primary strength – hiding the backdoor pattern by making it imperceptible to humans – against it, and propose to learn a perceptible (stronger) pattern (also during the training) that can overpower the attacker's imperceptible (weaker) pattern.  
We demonstrate the effectiveness of the proposed defensive mechanism through various commonly used replay-based (both generative and exact replay-based) continual learning algorithms using continual learning benchmark variants of CIFAR-10, CIFAR-100, and MNIST datasets. Most noteworthy, 
we show that our proposed defensive framework considerably improves the robustness of continual learning algorithms with no knowledge of the attacker's target task, attacker's target class, shape, size, and location of the attacker's pattern. Moreover, our defensive framework does not depend on the underlying continual learning algorithm and does not rely on detecting the attack samples and subsequently removing them from further consideration but it, instead,
attempts to correctly classify even the attack samples and thus ensuring robustness in continual learning models.
We term our defensive framework as \textit{Adversary Aware Continual Learning (AACL)}.
\end{abstract}


\titlepgskip=-21pt

\maketitle

\section{Introduction}
\label{sec:introduction}
\PARstart{C}{ontinual} learning (CL) represents a setting where a model is asked to learn from sequential data with evolving distributions \cite{de2019continual}. To be useful, a continual learning model has to strike a balance between two opposing characteristics: i) stability, which refers to model’s ability to retain previously learned but still relevant knowledge, and ii) exhibit plasticity, which refers to the model’s ability to acquire new knowledge and adapt itself to a possibly drifting or changing distributions. Traditionally the main challenge for CL models is to maintain stability, i.e. the CL models face difficulty in retaining the previously acquired knowledge while they are asked to learn new knowledge, a common phenomenon referred to as catastrophic forgetting \cite{mccloskey1989catastrophic}. Therefore, much of the work in continual
learning has focused on avoiding catastrophic forgetting while maintaining this delicate balance between stability and plasticity. 

While several CL approaches have been proposed to avoid the problem of catastrophic forgetting, but recently it has been found that these approaches are extremely vulnerable to adversarial backdoor attacks \cite{umer2021adversarial, umer2020targeted, umer2022false}, where an intelligent adversary can easily insert miniature amount of misinformation in the training data to deliberately or intentionally disturb the balance between stability and plasticity acquired by the CL model. More specifically, the goal of such an attack is to artificially increase the forgetting of the CL model on a explicitly targeted previously learned task. 

In this work, we propose a novel defensive framework to ensure robustness in CL models to imperceptible misinformation inserted via adversarial backdoor attack. Our defensive framework utilizes a small amount of defensive (decoy) samples to inhibit the impact of the malicious samples. The defensive (decoy) samples also contain a pattern similar to adversarial backdoor malicious samples but this pattern is: i) perceptible (stronger); and ii) different than the attacker's unknown imperceptible (weak) pattern.
We use the intensity of a pattern, i.e., perceptibility of a pattern to determine the strength of the pattern. The more stronger pattern is more perceptible and vice-versa. Specifically, as the malicious samples (falsely labeled samples containing attacker's imperceptible pattern) are appended to the training data by the adversary, we as a defender also provide additional defensive samples during training. 
The goal here is two-fold: i) to force the CL model to learn to not only correctly classify the clean samples of each task but also to correctly classify the defensive samples, i.e., the samples with the defensive pattern, ii) to weaken the impact of the attacker's imperceptible pattern in the presence of the defender's perceptible pattern. Once the training is done, the defensive pattern is applied to all the test samples at test time including those that contain the unknown attacker's malicious pattern. 
In the presence of stronger defensive pattern, the CL model ignores the weaker attacker's pattern and correctly classify all the test samples including the malicious ones.
In other words, the defensive samples serve to inoculate the CL model against the malicious samples. We referred to such learning as the \textit{Adversary Aware Continual Learning (AACL)}.

We note that on the surface AACL may seem similar to well known state of the art adversarial training (AT) \cite{goodfellow2014explaining} defense to defend against adversarial examples \cite{szegedy2013intriguing, papernot2016practical, papernot2016transferability}. However, in section 5 we compare AACL with adversarial training and show that AACL is not only different but also more efficient than adversarial training. Our paper is organized as follows; section 2 briefly describes the class-incremental learning and adversarial threats to class incremental learning, 
section 3 discusses the existing backdoor defenses and their limitations in class incremental learning setting, section 4 describes the proposed AACL defensive mechanism against the adversarial backdoor attacks to class incremental learning models, section 5 compares the AACL framework against the adversarial training, and section 6 constitutes the experiment and results followed by conclusions section.

\section{Class Incremental Learning \& Adversarial threats}
Continual (incremental) learning can be categorized into three different scenarios, which are i) Task incremental learning, ii) Domain incremental learning, and iii) Class incremental learning \cite{vandeven2019three}.
Out of these three different scenarios, class incremental learning (CIL) represents the most realistic, challenging, and practical continual learning (CL) scenario \cite{vandeven2019three}, where the CL model is asked to learn new classes from a sequence of different tasks. In CIL setting, the model does not have access to the task-ID at test time, therefore, the model not only needs to correctly predict the classes but also needs to infer from which task these classes are coming from. We will focus on class incremental learning in this work.
Class-incremental learning can also be utilized in various practical scenarios \cite{masana2020class} for instance; 1) in the situation where there is a privacy risk in storing the data such as patient's data in hospital setting \cite{de2019continual, mcclure2018distributed}; 2) reducing the computational cost of training deep learning models when the model is required to train every time it receives new data, and; 3) for those systems that have limitations on their memory and can not simply store even fixed small amount of data from all the previously encountered task(s) such as robotics or applications running on small portable smart devices like mobile phones. For all these applications, class incremental learning can play a vital role.

As mentioned previously, class incremental learning suffers from the problem of catastrophic forgetting \cite{mccloskey1989catastrophic}. The most successful CIL approaches that are proposed to avoid catastrophic forgetting are the replay-based approaches, which either i) store original data from the previous tasks, replay them with the data from the current task while optimizing network parameters jointly over essentially the data from all tasks or ii) use a generative model to generate pseudo-data to be replayed with the real data samples. 
Incremental task-agnostic metal learning (ITAML) \cite{rajasegaran2020itaml}, 
Deep Generative Replay (DGR) \cite{shin2017continual}, Deep Generative Replay with Distillation \cite{vandeven2018generative,vandeven2019three}, and Random Path Selection (RPS-net) \cite{rajasegaran2019random} are examples of replay-based approaches.

It has been shown recently that class incremental learning algorithms are highly vulnerable to adversarial backdoor attacks \cite{umer2022false}. Adversarial Backdoor attacks are specifically designed targeted poisoning attacks \cite{gu2017badnets,shafahi2018poison}, where a specific backdoor pattern (or tag) is used to cause intended misclassification. These attacks are insidious and are difficult to detect as the model will only give erroneous output in the presence of the malicious backdoor pattern (or tag); and otherwise perform normally on the clean data.  The adversarial backdoor attack's threat is even more pronounced in the continual learning setting as the malicious backdoor pattern can be provided to a small fraction of training data of any task of attacker's choice. In other words the attacker can pick any task of its own choice as the \textit{target task} and can adversely affect its performance at test time or can deliberately increase the forgetting of any task. Furthermore, the adversarial backdoor attacks to class incremental learning models are more stealthier as the malicious backdoor pattern is completely \textit{imperceptible} to human eye. Because of the imperceptibility of the backdoor pattern, the attacker can easily hide the malicious samples in any task of its own choice without being detected.  

\section{Conventional backdoor defenses \& their limitations in continual learning setting}
Existing backdoor defenses are designed to defend against backdoor attacks in conventional \textit{stationary} settings. Therefore, these approaches cannot be simply extended to continual / incremental settings. Existing backdoor defenses can be broadly categorized into three different categories; i) training time defenses, ii) inference-time defenses (during testing), and iii) model correction defenses. We briefly explain these defenses along with their limitations to succeed in continual / incremental learning settings.

\textit{Training-time defenses} assume that the defender has access to the training data. The goal of these defenses is to detect and remove the malicious samples. These defenses commonly utilize anomaly detection techniques. Examples of such defenses are spectral signatures \cite{tran2018spectral}, activation clustering \cite{chen2019detecting}, and gradient clustering \cite{chan2019poison}. In continual / incremental learning  setting, such defenses are not practical as anomaly detection is required at each time step, which is computationally very expensive. Furthermore, these defenses assume access to the training dataset that is potentially compromised. Such assumption is not practicable and impossible in continual setting because it is unknown apriori which task has been compromised.

\textit{Inference-time defenses} aim to detect and remove backdoor pattern at test/inference time. These defenses rely on the fact that the model will perform reasonably well on the samples that do not contain backdoor pattern. Therefore, such defenses usually do some pre-processing on the test samples before providing them to the model. For instance, STRIP \cite{gao2019strip} superimposes various patterns on the input test samples and expects that the predictions of the model would be random for clean inputs but more consistent for the input that contains the backdoor pattern. Neo  \cite{udeshi2022model} seeks to systematically search the location of the backdoor pattern and then modify the image by blocking the trigger. Li et al. \cite{li2020rethinking} apply spatial transformations on the test images to change the location of the backdoor pattern. Doan et al. \cite{doan2020februus} use GradCam \cite{selvaraju2016grad} to detect the presence of the backdoor triggers. Such defenses are not feasible in the class incremental setting as similar to the training time case, it is also not known a-priori which task contains the backdoor pattern at inference time. Such defenses need to be applied at each time step during testing, immensely increasing the computational cost. Moreover, pre-processing the images for clean tasks may degrade the test time performance of clean images.

\textit{Model correction defenses}, on the other hand, aim to correct the trained model for any backdoor vulnerabilities. For instance, fine pruning \cite{liu2018fine} assumes that neurons activated by clean inputs and backdoor inputs are different. Therefore, fine pruning sorts the neurons based on their activation on the clean inputs and prune those neurons that contribute least to the classification task at hand. Artificial brain stimulation (ABS) \cite{liu2019abs} scans neurons and use reverse-engineering techniques to generate backdoor pattern candidates. Backdoor suppression \cite{sarkar2020backdoor} builds a wrapper around the trained model to neutralize the effect of the backdoor pattern. Multiple noisy versions of an input are provided to the model and the final prediction is obtained by applying the majority vote on the multiple noisy replicas of the input. Neural cleanse \cite{wang2019neural} reverse engineers the backdoor pattern via optimization.

Model correction based approaches serve as a reasonable defense against the conventional backdoor attacks; however, these defenses have major shortcomings even in the conventional stationary setting. For instance, fine pruning \cite{liu2018fine} assumes that neurons are activated differently for clean and backdoor inputs, which is not true in practice and therefore, cannot be assumed. Also, pruning neurons degrades the clean accuracy of the model. Similarly, backdoor suppression \cite{sarkar2020backdoor} generates different noisy replicas of the input, which severely degrades the accuracy of the model on the clean inputs. The approaches that reverse engineer backdoor pattern such as ABS \cite{liu2019abs}  and neural cleanse \cite{wang2019neural} are computationally demanding. Moreover, they often do not reverse engineer a pattern that is reasonably similar to the one used by the attacker, resulting in very limited success only in select few scenarios. In continual / incremental setting, we cannot simply prune neurons as some of the neurons contain useful information for the previous task(s). Finally, as the number of tasks grow in incremental setting, all such defenses not only become more computationally expensive but also impractical.

\section{Adversary Aware Continual Learning (AACL)}
To counter the impact of the adversarial backdoor attack in class incremental learning (CIL) setting, we propose a novel defensive framework called adversary aware continual learning (AACL). AACL framework is inspired from the well known and robust adversarial training defense \cite{goodfellow2014explaining, bairecent} proposed to defend \textit{static} deep learning models against test (inference) time adversarial samples. Adversarial training is an intuitive defense that aims to improve the robustness of the deep learning models by training it with adversarial samples. Our proposed AACL framework also aims to improve the robustness of the CL model during training. However, the backdoor attack to a CL model happens during training, and the defender is not aware of the nature of the backdoor pattern ( e.g., shape, size or its location) chosen by the attacker. If the defender knows the nature of the backdoor pattern, the defender can simply search for the pattern in the training samples and thus can easily remove or detect these malicious samples during training. AACL framework, therefore, aims to train a CL model with samples that contain a different but stronger pattern to reduce the impact of the attacker's unknown backdoor pattern. We note that the attacker chooses its pattern to be imperceptible (to humans), with the intention of being stealthier. Indeed, an imperceptible attack is more difficult to detect and defend against. However, in our proposed approach we attempt to use the attacker's strength against it, by developing the defense specifically for imperceptible attack patterns. We show that the defender can easily use a \textit{perceptible} pattern as a stronger pattern to overpower the attacker's imperceptible (weaker) pattern. In our formulation, the goal of the defender is to force the CL model to weaken or mitigate the association of the attacker's pattern to the incorrect label in the presence of the defender's pattern. We refer to the defender's perceptible pattern as the \textit{defensive pattern}.

In AACL, the defender borrows from the attacker's playbook and also inserts an additional \textit{decoy} samples into the training data. However, these samples have the defender's \textit{defensive pattern} applied to them and are assigned the correct labels. We refer to these decoy training samples as the \textit{defensive samples}. 
Therefore, for any given task $t$, the CL model is trained with the original clean training samples, some amount of unknown malicious samples provided by the adversary, and a small number of defensive samples provided during that time step for task $t$. The goal, at the end of the training, is to have the CL model to learn to disassociate the attacker's pattern with the incorrect label in the presence of the defensive pattern.


At inference time, the defender provides the defensive pattern to \textit{all} test samples including, of course, the malicious samples whose identities it does not know. As the defensive pattern is stronger (more perceptible) than the attacker's pattern, our expectation of the CL model is to put more focus to the defensive pattern while making its decision, ignore the weaker (imperceptible) attacker's pattern, and ultimately make the correct classification on the malicious samples. In other words, the defensive samples serve to \textit{inoculate} the CL model against the malicious samples.  

Note that the defender is unaware of the attacker's target task, the attacked data or the attacker's targeted label. Therefore, the defender provides the defensive pattern to a small number of samples of each class of each task during training time (and to all samples at inference time), which additionally helps the CL model to learn to correctly classify the samples with the defensive pattern. In other words, unbeknownst to the defender, the attacker can insert malicious samples into any task(s) of its choice and thus has an arguably unfair advantage over the defender. However, we show that -- despite adversary's advantage -- our Adversary Aware Continual Learning (AACL) framework reasonably improves the robustness and accuracy of the CL model.

\subsection{AACL framework for Replay-based CL approaches}
We now formally describe the AACL framework. Mathematically, if $\mathcal{X}^t$ denotes the  training data at the current time-step with their corresponding true labels $\mathcal{Y}^t$, replay-based approaches minimize the following generalized loss function
\begin{equation}
\label{eq_dgr}
\begin{split}
    \mathcal{L}(\mathcal{F}_{\theta}) = &  \mathcal{L}_{current}[\mathcal{F}_{\theta}(\mathcal{X}^t),\mathcal{Y}^t] + \\ &  \mathcal{L}_{replay}[\mathcal{F}_{\theta}(\mathcal{X}^{t-1} \cup \mathcal{X}^{t-2} \cup \dots \mathcal{X}^1), \\ & \hspace{1.1in} \mathcal{Y}^{t-1} \cup \mathcal{Y}^{t-2} \cup \dots \mathcal{Y}^1]
\end{split}
\end{equation}
where,  $\mathcal{L}_{current}[\mathcal{F}_{\theta}(\mathcal{X}^t),\mathcal{Y}^t]$ is the loss on current data, and $\mathcal{L}_{replay}[\mathcal{F}_{\theta}(\mathcal{X}^{t-1} \cup \mathcal{X}^{t-2} \cup \dots \mathcal{X}^1), \mathcal{Y}^{t-1} \cup \mathcal{Y}^{t-2} \cup \dots \mathcal{Y}^1]$ is the loss on the data replayed from all previous tasks. Here, $\mathcal{X}^k$, $k = t-1,t-2, \dots, 1$ are the replayed samples for all the previous tasks and $\mathcal{Y}^k$ are their corresponding correct labels.

To attack the replay-based approaches, while training on the current task, the adversary appends a small amount of additional malicious samples that contain the imperceptible (weak) attacker's pattern to the training data of the current task. Mathematically, let $\mathcal{X}^{t}_{b}$ represent the small amount of malicious backdoor samples inserted in the training data of the current task, and $\mathcal{Y}^{t}_{b}$ be their corresponding \textit{false} labels (attacker's desired target labels). The loss function that replay-based approaches will minimize with the backdoor samples is then:
\begin{equation}
\label{eq_dgr_w_bd}
\begin{split}
    \mathcal{L}(\mathcal{F}_{\theta}) = &  \mathcal{L}_{current}[\mathcal{F}_{\theta}(\mathcal{X}^t \cup \mathcal{X}^{t}_{b}),(\mathcal{Y}^t \cup \mathcal{Y}_{b}^{t})] + \\ & \mathcal{L}_{replay}[\mathcal{F}_{\theta}(\mathcal{X}^{t-1} \cup \mathcal{X}^{t-2} \cup \dots \mathcal{X}^1), \\ & \mathcal{Y}^{t-1} \cup \mathcal{Y}^{t-2} \cup \dots \mathcal{Y}^1]
\end{split}
\end{equation}

To defend the replay-based approaches, AACL framework provides small amount of additional training samples, which we term as defensive (decoy) samples that contain the perceptible (strong) defensive pattern in to the training data of the current task. 
It is important to mention here that to counter such an attack, the defender ideally needs to provide the defensive pattern to the classes of the attacker's targeted task. However the attacker's target task and the attacker's targeted label is \textbf{NOT} known to the defender. Hence, during the training on the current task, the defender provides the defensive pattern to all the classes of the current task, and to all the classes of each previous task(s). For this, the defender picks a small fixed amount of samples from each class of each task (including current and previous task(s)), insert the perceptible defensive pattern to these samples, assign them the \textbf{correct} label, and append them with the training data of the current task. In other words, the defender is providing the defensive (decoy) samples to the current task, but also replaying the defensive (decoy) samples from each previous task with the training data of the current task. The goal with this training is to force the model to correctly classify the defensive samples that contain the defensive pattern. At test (inference) time, the defensive pattern is provided to all the test samples. When the model sees the test sample with both the attacker's imperceptible (weak) pattern and the defender's perceptible (strong) pattern, the model makes the decision based on the presence of the stronger defensive pattern along with the genuine features of the image, and give the correct classification of the test sample.



Mathematically, let $\mathcal{X}^{k}_{d}$ represent the total amount of defensive samples inserted in the training data of the $k^{th}$ task and $\mathcal{Y}^{k}_{d}$ be their corresponding \textit{correct} labels. The loss function that our proposed AACL framework minimizes is as follows:
\begin{equation}
\label{eq_dgr_w_bd_defense}
\begin{split}
    \mathcal{L}(\mathcal{F}_{\theta}) = &  \mathcal{L}_{current}[\mathcal{F}_{\theta}(\mathcal{X}^t \cup \mathcal{X}^{t}_{b}\cup \mathcal{X}^{t}_{d}),(\mathcal{Y}^t \cup \mathcal{Y}_{b}^{t} \cup \mathcal{Y}^{t}_d)] + \\ & \mathcal{L}_{replay}[\mathcal{F}_{\theta}((\mathcal{X}^{t-1} \cup \mathcal{X}^{t-1}_{d}) \cup (\mathcal{X}^{t-2}\cup \mathcal{X}^{t-2}_{d}) \cup \dots \\ & (\mathcal{X}^1 \cup \mathcal{X}^{1}_{d})), 
   (\mathcal{Y}^{t-1} \cup \mathcal{Y}^{t-1}_d) \cup (\mathcal{Y}^{t-2}\cup \mathcal{Y}^{t-2}_d) \cup \dots \\ &(\mathcal{Y}^1 \cup \mathcal{Y}^{1}_d)]
\end{split}
\end{equation} 


The generalized pseudo-code for the original replay-based approaches are shown in Algorithm \ref{AACL_er_orig} for exact replay-based approaches and in Algorithm \ref{AACL_gr_orig} for generative replay-based approaches respectively.

The generalized pseudo-code of Adversary Aware Continual Learning (AACL) framework for defending replay-based CL approaches is shown in Algorithm \ref{AACL_er_def} for exact replay-based approaches, and in Algorithm \ref{AACL_gr_def} for generative replay-based approaches respectively. 

\begin{algorithm}
\caption{Exact Replay-Based CL Approaches}\label{AACL_er_orig}
\hspace*{\algorithmicindent} \textbf{Input} $(\mathcal{X}^t,\mathcal{Y}^t)$: Training data samples received for time step $t$; $T$: total number of tasks; $\mathcal{F}_{\theta}$: Initial model parametrized by $\theta$; 
\\
\hspace*{\algorithmicindent} \textbf{Output} Optimal Parameters $\theta^*$ and the final model $\mathcal{F}_{\theta^*}$
\begin{algorithmic}[1]

    \For{$t= 1,...,T$:}
    \If {$t == 1$}
    \State Use the original training data from the first time step
    \Else
    \For{k = 1,...,t-1}
    \State Pick a fixed amount of informative samples from the previous task(s), i.e., ($\mathcal{X}^{k}, \mathcal{Y}^{k}$) data-label pair to be replayed with the current task's training data
    \EndFor
    \EndIf
    \State $\theta_t^* \leftarrow \displaystyle\minimize_{\theta} \mathcal{L}[\mathcal{F}_{\theta}(\mathcal{X}^t),\mathcal{Y}^t] +  \mathcal{L}[\mathcal{F}_{\theta}
    (\mathcal{X}^{t-1} \cup \mathcal{X}^{t-2} \cup \dots \mathcal{X}^1), 
    (\mathcal{Y}^{t-1}\cup \mathcal{Y}^{t-2}\cup   \dots  \mathcal{Y}^1)]$)
    \EndFor
\end{algorithmic}
\end{algorithm}

\begin{algorithm}
\caption{Adversary Aware Continual Learning Framework For Exact Replay-Based CL Approaches}\label{AACL_er_def}
\hspace*{\algorithmicindent} \textbf{Input} $(\mathcal{X}^t,\mathcal{Y}^t)$: Training data samples received for time step $t$; $T$: total number of tasks; $\mathcal{F}_{\theta}$: Initial model parametrized by $\theta$; 
\\
\hspace*{\algorithmicindent} \textbf{Output} Optimal Parameters $\theta^*$ and the final model $\mathcal{F}_{\theta^*}$
\begin{algorithmic}[1]

    \For{$t= 1,...,T$:}
    \If {$t == 1$}
    \State Use the original training data from the first time step
    \Else
    \For{k = 1,...,t-1}
     \State Pick a fixed amount of informative samples from the previous task(s), i.e., ($\mathcal{X}^{k}, \mathcal{Y}^{k}$) data-label pair to be replayed with the current task's training data
    \State Pick a small fixed amount of samples from each class of previous task(s) and add perceptible defensive pattern to these samples to create defensive samples $(\mathcal{X}^{k}_{d}, \mathcal{Y}^{k}_{d})$
    \State Append defensive samples to the 
    training data of the previous task, i.e., Append $(\mathcal{X}^{k}_{d}, \mathcal{Y}^{k}_{d})$ to $(\mathcal{X}^k, \mathcal{Y}^k)$ 
    \EndFor
    \EndIf
    \State Malicious backdoor samples $(\mathcal{X}^t_{b}, \mathcal{Y}_{b}^t)$ unbeknownst to the defender, are provided in the training data at the current time-step $t$ by an adversary
    \State Append defensive samples to the 
    training data of the current task, i.e., Append $(\mathcal{X}^{t}_{d}, \mathcal{Y}^{t}_{d})$ to $(\mathcal{X}^t, \mathcal{Y}^t)$ 
    \State $\theta_t^* \leftarrow \displaystyle\minimize_{\theta}\hspace{0.1in} \mathcal{L}[\mathcal{F}_{\theta}(\mathcal{X}^t \cup \mathcal{X}^{t}_{b}\cup \mathcal{X}^{t}_{d}),(\mathcal{Y}^t \cup \mathcal{Y}_{b}^{t}\cup \mathcal{Y}_{d}^{t})] +  \mathcal{L}[\mathcal{F}_{\theta}((\mathcal{X}^{t-1} \cup \mathcal{X}^{t-1}_{d}) \cup (\mathcal{X}^{t-2}\cup \mathcal{X}^{t-2}_{d}) \cup \dots  $\par$ (\mathcal{X}^1 \cup \mathcal{X}^{1}_{d})), (\mathcal{Y}^{t-1} \cup \mathcal{Y}^{t-1}_d) \cup (\mathcal{Y}^{t-2}\cup \mathcal{Y}^{t-2}_d) \cup   \dots $\par$ (\mathcal{Y}^1 \cup \mathcal{Y}^{1}_d)]$)
    \EndFor
\end{algorithmic}
\end{algorithm}

\begin{algorithm}
\caption{Generative replay-based CL Approaches}\label{AACL_gr_orig}
\hspace*{\algorithmicindent} \textbf{Input} $(\mathcal{X}^t,\mathcal{Y}^t)$: Training data samples received for time step $t$; $T$: total number of tasks; $\mathcal{F}_{\theta}$: Initial model parameterized by $\theta$; $\mathcal{G}_{\phi}$: Initial generator parameterized by $\phi$ \\
\hspace*{\algorithmicindent} \textbf{Output} Optimal Parameter $\theta^*$ and the final model $\mathcal{F}_{\theta^*}$
\begin{algorithmic}[1]

    \For{$t= 1,...,T$:}
    \If {$t == 1$}
    \State Use the original training data from the first time step
    \Else
    \For{k = 1,...,t-1}
    \State Generate samples from the previous task(s), i.e., $\mathcal{X}^k \sim \mathcal{G}_{\phi}$
    \State Label generated samples from the previous optimal model, i.e., $\Tilde{\mathcal{Y}}^k \leftarrow \mathcal{F}_{\theta_{t-1}^*}(\mathcal{X}^k)$
    \EndFor
    \EndIf
    \State $\theta_t^* \leftarrow \displaystyle{\minimize_{\theta}(\mathcal{L}[\mathcal{F}_{\theta}(\mathcal{X}^t),\mathcal{Y}^t]}\hspace{0.1in} + 
    (\mathcal{L}[\mathcal{F}_{\theta}(\mathcal{X}^{t-1} \cup \mathcal{X}^{t-2} \cup ...... \mathcal{X}^1), 
    (\Tilde{\mathcal{Y}}^{t-1} \cup \Tilde{\mathcal{Y}}^{t-2} \cup ...... \Tilde{\mathcal{Y}}^1)] $)
    \State $\phi_t^* \leftarrow \displaystyle{\minimize_{\phi}(\mathcal{L}[\mathcal{G}_{\phi}(\mathcal{X}^t)]} +  (\mathcal{L}[\mathcal{G}_{\phi}(\mathcal{X}^{t-1} \cup \mathcal{X}^{t-2} \cup ...... \mathcal{X}^1)] $)
    \EndFor
\end{algorithmic}
\end{algorithm}

\begin{algorithm}
\caption{Adversary Aware Continual Learning Framework For Generative Replay-Based CL Approaches}\label{AACL_gr_def}
\hspace*{\algorithmicindent} \textbf{Input} $(\mathcal{X}^t,\mathcal{Y}^t)$: Training data samples received for time step $t$; $T$: total number of tasks; $\mathcal{F}_{\theta}$: Initial model parameterized by $\theta$; $\mathcal{G}_{\phi}$: Initial generator parameterized by $\phi$ 
\\
\hspace*{\algorithmicindent} \textbf{Output} Optimal Parameters $\theta^*$ and the final model $\mathcal{F}_{\theta^*}$
\begin{algorithmic}[1]

    \For{$t= 1,...,T$:}
    \If {$t == 1$}
    \State Use the original training data from the first time step
    \Else
    \For{k = 1,...,t-1}
    \State Generate samples from the previous task, i.e., $\mathcal{X}^k \sim \mathcal{G}_{\phi}$
    \State Label generated samples from the previous optimal model, i.e., $\Tilde{\mathcal{Y}}^k \leftarrow \mathcal{F}_{\theta_{t-1}^*}(\mathcal{X}^k)$
    \State  Pick a small fixed amount of samples from each class of previous task(s) and add perceptible defensive pattern to these samples to create defensive samples $(\mathcal{X}^{k}_{d}, \mathcal{Y}^{k}_{d})$
    \State Append defensive samples to the 
           training data of the previous task, i.e., Append $(\mathcal{X}^{k}_{d}, \mathcal{Y}^{k}_{d})$ to $(\mathcal{X}^k, \Tilde{\mathcal{Y}}^k)$ 
    \EndFor
    \EndIf
    \State Malicious backdoor samples $(\mathcal{X}^t_{b}, \mathcal{Y}_{b}^t)$ unbeknownst to the defender, are provided in the training data at the current time-step $t$ by an adversary
    \State Append defensive samples to the 
    training data of the current task, i.e., Append $(\mathcal{X}^{t}_{d}, \mathcal{Y}^{t}_{d})$ to $(\mathcal{X}^t, \mathcal{Y}^t)$ 
    \State $\theta_t^* \leftarrow \displaystyle\minimize_{\theta}\hspace{0.1in} \mathcal{L}[\mathcal{F}_{\theta}(\mathcal{X}^t \cup \mathcal{X}^{t}_{b}\cup \mathcal{X}^{t}_{d}),(\mathcal{Y}^t \cup \mathcal{Y}_{b}^{t}\cup \mathcal{Y}_{d}^{t})] +  \mathcal{L}[\mathcal{F}_{\theta}((\mathcal{X}^{t-1} \cup \mathcal{X}^{t-1}_{d}) \cup (\mathcal{X}^{t-2}\cup \mathcal{X}^{t-2}_{d}) \cup \dots  $\par$ (\mathcal{X}^1 \cup \mathcal{X}^{1}_{d})), (\Tilde{\mathcal{Y}}^{t-1} \cup \mathcal{Y}^{t-1}_d) \cup (\Tilde{\mathcal{Y}}^{t-2}\cup \mathcal{Y}^{t-2}_d) \cup   \dots $\par$ (\Tilde{\mathcal{Y}}^1 \cup \mathcal{Y}^{1}_d)]$)
    \State $\phi_t^* \leftarrow \displaystyle{\minimize_{\phi} \hspace{0.1in}(\mathcal{L}[\mathcal{G}_{\phi}(\mathcal{X}^t)]} +  (\mathcal{L}[\mathcal{G}_{\phi}(\mathcal{X}^{t-1} \cup \mathcal{X}^{t-2} \cup ...... \mathcal{X}^1)] $)
    \EndFor
\end{algorithmic}
\end{algorithm}

\section{Comparison of Adversary Aware Continual Learning framework to Adversarial Training}
Adversarial training is one of the most popular
and reasonably robust proposed defenses against adversarial examples \cite{szegedy2013intriguing, zhang2019adversarial, hendrycks2021natural}. Adversarial examples are malicious samples well-known for evading deep learning models at test time. Adversarial example is generated by adding a strategically chosen imperceptible perturbation to the test sample. Mathematically, a sample adversarial example is generated by maximizing the following objective function at the test time
\begin{equation}
    \Max_{||\delta||<\epsilon}(\mathcal{L}(\mathcal{F}_{\theta}(x + \delta), y))
\label{adv_ex}    
\end{equation} 
where $\mathcal{L}$ in equation \ref{adv_ex} denotes the loss function, for instance cross-entropy loss \cite{zhang2018generalized, martinez2018taming}. For a clean test sample $x_{test}$, equation \ref{adv_ex} finds the perturbation $\delta$ within some norm bound $\epsilon$, such that when the perturbation is added to the test sample, the loss $\mathcal{L}$ is maximized. To defend against adversarial examples at test time, adversarial training proposes to solve the following min-max objective function during training  
\begin{equation}
    \Min_\theta(\frac{1}{N} \sum_{i=1}^N  \Max_{||\delta||<\epsilon}(\mathcal{L}(\mathcal{F}_{\theta}(x_i + \delta), y_i))
\label{adv_tr}  
\end{equation}
Where $N$ is the number of training examples. In other words, the model with adversarial training scheme is not learning the clean training samples but rather the perturbed or adversarial version of the clean samples. Once the model is trained with the adversarial examples, the model is considered to be robust against adversarial examples. 
Equation \ref{adv_tr} can also be re-written as follows
\begin{equation}
    \Min_\theta(\frac{1}{N} \sum_{i=1}^N  (\mathcal{L}(\mathcal{F}_{\theta}(x_i + \delta^*), y_i))
\label{adv_tr_new}  
\end{equation}
where, $\delta^* = argmax_{||\delta||<\epsilon}(\mathcal{L}(\mathcal{F}_{\theta}(x_i + \delta), y_i)$. More specifically, adversarial training first tries to find the optimal perturbations $\delta^*$ within some norm bound for a particular example and then minimizes the loss function on that example.

The malicious backdoor sample can also be considered as a sample generated through adding an imperceptible perturbation $\delta_{att}$ to a randomly picked training sample $x$, i.e.,  $x_b = x + \delta_{att}$. In our case, the imperceptible perturbation $\delta_{att}$ refers to the imperceptible backdoor pattern. This malicious sample is added to the training data with the attacker's chosen false label $y_b$. On the other hand, our defensive sample is generated through adding the defender's chosen optimal (perceptible) perturbation $\delta^*_{def}$ to a randomly picked clean training sample, i.e., $x_d = x + \delta^*_{def}$ in the form of the defensive pattern. This defensive sample is added to the training data with its true label $y$. The goal here is to force the CL model to learn to correctly classify the defensive sample with the optimal perturbation $\delta^*_{def}$.
We refer to the defender's chosen perturbation as the \textit{optimal} perturbation as it represents the stronger (perceptible) noise that can be added to the sample relative to the attacker's weaker (imperceptible) noise. Once the model learns to disassociate the attacker's weaker noise in the presence of the relatively stronger defensive noise, the model correctly classifies the malicious sample that contains both attacker's noise (imperceptible) and the defender's noise (perceptible).

Mathematically, if at a particular time step during the training of the CL model, there are $N$ original training samples, $B$ malicious samples, and $D$ defensive samples, the CL model with our proposed Adversary Aware Continual Learning (AACL) framework will then be minimizing the following objective function

\begin{equation}
\begin{split}
    \Min_\theta ( \frac{1}{N+B+D} (\sum_{i=1}^N \mathcal{L}(\mathcal{F}_{\theta}(x_i), y_i) + 
    \sum_{j=1}^B \mathcal{L}(\mathcal{F}_{\theta}(x_j + \delta_{att}),
    \\
    y^b_j) +
    \sum_{k=1}^D \mathcal{L}(\mathcal{F}_{\theta}(x_k + \delta^*_{def}), y_k)))   
\end{split}
\label{LF_w_bd_CL}
\end{equation}

The expression $\sum_{k=1}^D \mathcal{L}(\mathcal{F}_{\theta}(x_k + \delta^*_{def}), y_k)$ in equation \ref{LF_w_bd_CL} looks similar to the loss function that adversarial training is trying to minimize in equation \ref{adv_tr_new}. However, the adversarial aware learning framework proposed against imperceptible backdoor training attacks has the following major differences than adversarial training defense proposed against test time adversarial examples: 

\begin{itemize}
    \item Adversarial training aims to learn -- during training -- the same or similar perturbations that the attacker wants to add at test time. We argue that this generally impractical, as the defender (in CL setting or otherwise) does not usually have the luxury to know the attacker's pattern (perturbations) except perhaps that the pattern may or may not be imperceptible. IN our formulation, the CL defender aims to learn a completely different but a stronger (perceptible) defensive pattern (perturbations) during training.
    \item In adversarial training, the defender provides adversarial perturbations to either all or some fixed (high) proportion of the training samples because the defender knows that there is no attack during training, the attack only happens at test time. In other words, the adversarial training assumes that all of the training samples are correctly labeled, which is not a good assumption to make, as there is always a chance of having few mislabeled samples in the training data \cite{song2022learning, natarajan2013learning, dawson2021rethinking}. We, as a CL defender on the other hand only add defensive pattern to a small amount of clean (correctly labeled) training samples per class per task at a particular time-step. 
    
    \item Adversarial training is known to be computationally ineffective as it needs to learn every possible perturbation that exists in the perturbation space \cite{zhang2019you, shafahi2019adversarial, wong2020fast}. Our adversarial aware continual learning framework, however, does not aim to learn the attacker's exact unknown imperceptible pattern (perturbations), it aims to learn an entirely different but fixed and stronger (perceptible) pattern during training through correctly labeled defensive samples. The stronger (perceptible) defensive pattern when presented with the attacker's weak (imperceptible) pattern in the test sample at the same time, the CL model ignores the attacker's pattern and thus provides the correct prediction.
\end{itemize}

In summary, we can say that our proposed defensive framework is \textit{inspired } by the adversarial training framework, but our proposed framework is more efficient, realistic and practical as compared to adversarial training.

\section{Experiments \& Results}
We evaluate our proposed adversary aware continual learning (AACL) defensive framework to defend various reply-based class incremental learning approaches against the adversarial imperceptible backdoor attacks. More specifically, we consider Deep Generative Replay (DGR) \cite{shin2017continual}, and Deep Generative Replay with Distillation \cite{vandeven2018generative,vandeven2019three} as the examples of generative replay-based class incremental learning algorithms, while Random Path Selection (RPS-net) \cite{rajasegaran2019random}, and Incremental task-agnostic metal learning (ITAML) \cite{rajasegaran2020itaml} as exact replay-based approaches. We consider the commonly continual variant of CIFAR-10, CIFAR-100, and MNIST datasets. Continual variants of CIFAR-10 and MNIST datasets contain 5 different tasks, where each task represents a binary classification problem. Continual version of CIFAR-100 dataset on the other hand represents a more challenging dataset, which constitutes 10 different tasks, where each task represents a 10-class classification problem.
\subsection{Defending Exact Replay-Based Class Incremental Learning Approaches using AACL}
\begin{figure}[!t]
\centering
\subfloat[]{\includegraphics[width=1.5in]{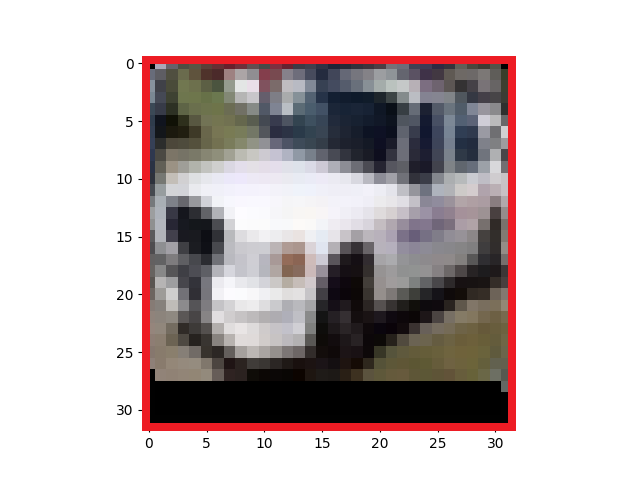}%
\label{img_att_patt_cifar10}}
\hfil
\subfloat[]{\includegraphics[width=1.7in]{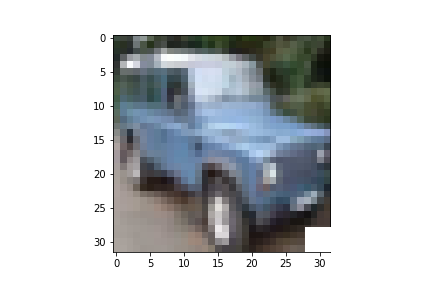}%
\label{img_def_patt_cifar10}}
\hfil
\caption{Imperceptible attack pattern \& Perceptible defense pattern for cifar-10 images: (a) image containing imperceptible frame as an attacker's pattern; (b) image containing perceptible square pattern as a defender's  pattern.}
\label{fig:mnist_inv_bd}
\end{figure}

When attacking CIFAR-10 dataset, the attacker can pick any of the five tasks of the continual learning variant of CIFAR-10 dataset as its desired target task. For now, we assume that the attacker's target task is Task 1, the attacker inserts a small amount (1\%) of malicious samples in to the training data of the target task. The attacker's desired false label is class 0 from Task 1. The attacker's pattern is an imperceptible rectangular frame of width 1, inserted around the boundary of the image as shown in Fig. \ref{img_att_patt_cifar10}. As the attack's pattern is imperceptible, we highlight the area in red in Fig. \ref{img_att_patt_cifar10}, where the attacker inserts it's imperceptible pattern. 

\begin{figure}[!t]
\centering
\subfloat[]{\includegraphics[width=1.1in]{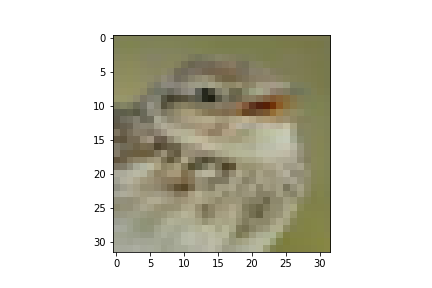}%
\label{img_org_cifar_f}}
\hfil
\subfloat[]{\includegraphics[width=1.1in]{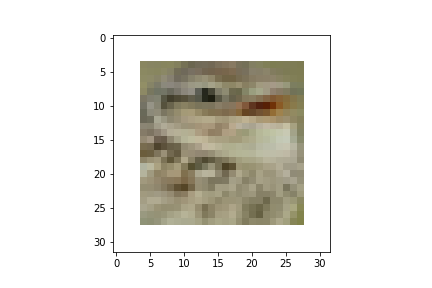}%
\label{img_vis_fr_cifar_f}}
\hfil
\subfloat[]{\includegraphics[width=1.1in]{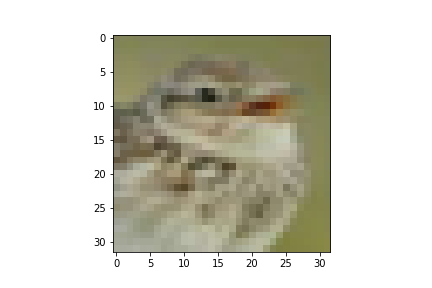}%
\label{img_inv_fr_wm_cifar_f}}
\hfil
\subfloat[]{\includegraphics[width=1.1in]{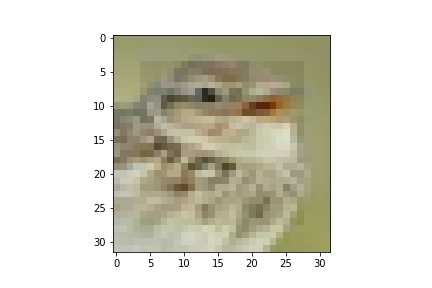}%
\label{img_vis_fr_wm_cifar_f}}
\caption{Sample CIFAR-10 images: (a) original image; (b) framed image; (c) image containing \textit{imperceptible} frame as a backdoor pattern with $\epsilon$ of 0.01; (d) image containing \textit{perceptible} frame as a backdoor pattern with $\epsilon$ of 0.1.}
\label{fig:cifar_inv_bd_f}
\end{figure}

To generate an imperceptible pattern for CIFAR-10 and CIFAR-100 images, we use a frame of pixels in the image that are imperceptible to human eye. To do so, we set $r_f$ to the original image with a frame whose values are set to one, and insert a backdoor pattern to the image as the weighted sum of the clean image $x$ and the framed image $r_f$. The weight of $r_f$ is set to $\epsilon$ and the weight of clean image $x$ is set to $1-\epsilon$ to obtain $x_{m} =(1-\epsilon) * x + \epsilon * r_f$. The imperceptibility of the backdoor pattern is controlled by $\epsilon$: smaller values make the pattern less noticeable to humans. A value of $\epsilon = 0.01$ results in a very imperceptible backdoor pattern. We use $\epsilon = 0.01$ in our experiments to make the backdoor pattern completely imperceptible to human eye. For visual purposes, a sample clean image, framed image, an image with an invisible backdoor pattern ($\epsilon=0.01$),  as well as a visible pattern  (with $\epsilon=0.1$) are shown in Figures \ref{img_org_cifar_f} , \ref{img_vis_fr_cifar_f}, \ref{img_inv_fr_wm_cifar_f}, and \ref{img_vis_fr_wm_cifar_f}, respectively for CIFAR-10 dataset. 


To make the CL model robust to such an insidious attack, the defender also provides  small amount of additional defensive samples in to the training data of the current task.
As the attacker's target task and target class is unknown to the defender, the defensive samples should cover all possible classes seen by the CL model so far until the current time step including the classes seen in the previous task(s). More specifically, for CIFAR-10 dataset, the defender additionally inserts 500 clean (correctly labeled) defensive samples from each class of each task (including both previous and current tasks) in to the training data at current time step. Note that the original CIFAR-10 dataset (without any malicious and defensive samples) contains 5000 samples per class per task. The defender provides its chosen defensive pattern to each defensive sample, which is perceptible (stronger) and entirely different than the attacker's unknown imperceptible (weaker) pattern. In particular, the defensive pattern is a white (strong intensity) square pattern provided at the bottom right corner of the defensive sample as shown in Fig. \ref{img_def_patt_cifar10}.

The model is then trained on a set of training data containing original clean samples, attacker's unknown malicious samples, and the defensive samples during the training of first task (attacker's target task), and trained on a set of clean and defensive samples for the remaining four tasks (untargeted tasks). Once the CL model learns to correctly classify the defensive samples containing  stronger defensive pattern, the same defensive pattern is then applied to all test samples at inference time. Those samples that contain the attacker's weaker imperceptible pattern when presented to the CL model along with the stronger defensive pattern, the latter overpowers the former, allowing the CL model to disassociate the image content with the imperceptible pattern, and hence mitigate the misclassification caused by the attack. 

Table \ref{tab:cifar10_exact_replay} shows the individual task's test time performance for both of the exact replay-based continual learning approaches, i.e., ITAML and RPS-net evaluated on the CIFAR-10 dataset. Results are presented with the mean and standard deviation computed over 5 independent runs for three different settings; i) clean (no attack setting), ii) attack (with attack only), and iii) defense (with attack and defense setting). The results show that our proposed adversary aware continual learning (AACL) defensive framework considerably improves the accuracy of the CL model for both of the algorithms. The defense performance on attacker's target task (Task 1) increases from 48.97\% to 89.71\%- an increase of about 41\% - for the ITAML algorithm. For RPS-net, the accuracy on the target task increases from 33.09\% to 63.53\% giving an increase of about 30\%.

CIFAR-100 represents a more challenging dataset, which has 10 different tasks in its  continual learning variant: each task is a 10-class classification problem. During training, the CL model is trained incrementally on these 10 different tasks, one task at a time. Once the training is complete, the CL model's goal is to correctly predict all 100 classes from these 10 tasks.
As before, the attacker can pick any of the ten tasks as its desired target task; without any loss of generality, we assume that the attacker's target task is Task 1, and the attacker's desired target label is class 7 from Task 1. In other words, the attacker attaches its invisible attack pattern to non-class-7 images and then assigns the label "class 7". The attacker inserts 1\% such malicious samples into the training data of Task 1. Similar to CIFAR-10, the attacker's pattern is the same imperceptible rectangular frame of width 1, inserted in the perimeter of the image. As the attacker's target task and class are not known to the defender, the defender attaches its defensive pattern to some small amount of defensive samples spanning all possible classes of each task seen thus far including both the current and previous task(s). In our experiment, for CIFAR-100 dataset, the defender provides 100 additional correctly labeled defensive samples per class per task. Note that the original CIFAR-100 dataset (without any malicious and defensive samples) contains 500 samples per class per task.

\begin{table*}[t]
    \centering
    \caption{
        Test accuracy (in \%) of Exact Replay-based class incremental learning approaches on CIFAR-10. Performances on Task 1 is in bold
    }
    \begin{tabular}{lcccccc}
        \multicolumn{7}{c}{} \\ \toprule
            & \multicolumn{3}{c}{ITAML}   & \multicolumn{3}{c}{RPS-net} \\ \cmidrule(lr){2-4} \cmidrule(lr){5-7}
        Tasks &   Clean   & Attack    & AACL Defense    & Clean & Attack   & AACL Defense     \\ \cmidrule(lr) {1-1} \cmidrule(lr){2-4} \cmidrule(lr){5-7}
        Task 1                    & $\mathbf{98.01} \pm \mathbf{0.36}$ & $\mathbf{48.97} \pm \mathbf{0.23}$  & $\mathbf{89.71} \pm \mathbf{1.69}$ & $\mathbf{62.87} \pm \mathbf{5.22}$ & $\mathbf{33.09} \pm \mathbf{3.17}$  & $\mathbf{63.53} \pm \mathbf{3.71}$   \\
        Task 2                 & $89.44 \pm 0.71$ & $88.60 \pm 0.38$ & $82.30 \pm 1.47$  & $45.48 \pm 2.49$ & $46.41 \pm 5.56$ & $48.30 \pm 2.39$  \\
        Task 3                    & $93.61 \pm 0.46$ & $92.58 \pm 0.24$  & $87.90 \pm 1.06$  & $74.63 \pm 4.91$ & $75.37 \pm 3.56$  & $61.69 \pm 1.24$    \\ 
        Task 4                    & $97.60 \pm 0.27$ & $97.07 \pm 0.27$  & $94.26 \pm 0.46$  & $86.19 \pm 1.77$ & $86.72 \pm 5.32$  & $80.84 \pm 1.80$ \\
        Task 5                    & $97.60 \pm 0.11$ & $96.69 \pm 0.28$  & $95.35 \pm 0.28$  & $84.46 \pm 3.78$ & $82.22 \pm 3.97$  & $90.06 \pm 1.20$ \\
        \bottomrule \\
    \end{tabular}
    \label{tab:cifar10_exact_replay}
\end{table*}

The individual task's test time performance for ITAML and RPS-net with CIFAR-100 dataset is shown in Table \ref{tab:cifar100_exact_replay}. It can be seen from Table \ref{tab:cifar100_exact_replay} that even for the challenging CIFAR-100 dataset, our proposed AACL defensive framework considerably improves the robust (defense) accuracy for both of the exact replay-based class incremental learning algorithms. For ITAML, the accuracy improves from about 10\% (Attack setting) to about 44\% (AACL defense setting) and for RPS-net, the accuracy improves from about 3\% (Attack setting) to about 30\% (AACL defense setting) for the attacker's desired target, i.e., Task 1. For CIFAR-100 dataset, we also observe a trade-off between natural accuracy on clean examples (examples from clean untargeted tasks, i.e., Task 2 to Task 10 in both clean and attack setting) and their corresponding AACL defense or robust accuracy, a common phenomenon observed in the literature for adversarial training based defenses as well \cite{raghunathan2019adversarial, tsipras2019robustness, wang2020once, zhang2019theoretically}.

\begin{table*}[t]
    \centering
    \caption{
        Test accuracy (in \%) of Exact Replay-based class incremental learning approaches on CIFAR-100. Performances on Task 1 is in bold
    }
    \begin{tabular}{lcccccc}
        \multicolumn{7}{c}{} \\ \toprule
            & \multicolumn{3}{c}{ITAML}   & \multicolumn{3}{c}{RPS-net} \\ \cmidrule(lr){2-4} \cmidrule(lr){5-7}
        Tasks &   Clean   & Attack    & AACL Defense    & Clean & Attack   & AACL Defense     \\ \cmidrule(lr) {1-1} \cmidrule(lr){2-4} \cmidrule(lr){5-7}
        Task 1                    & $\mathbf{80.62} \pm \mathbf{0.51}$ & $\mathbf{9.88} \pm \mathbf{1.46}$  & $\mathbf{44.14} \pm \mathbf{2.59}$ & $\mathbf{34.34} \pm \mathbf{3.78}$ & $\mathbf{ 2.94} \pm \mathbf{0.13}$  & $\mathbf{30.38} \pm \mathbf{0.43}$   \\
        Task 2                 & $76.28 \pm 0.94$ & $75.80 \pm 0.77$ & $60.04 \pm 1.07$  & $29.08 \pm 4.06$ & $27.86 \pm 0.40$ & $28.10\pm 0.49$  \\
        Task 3                    & $76.76 \pm 0.89$ & $76.68 \pm 0.52$  & $65.30 \pm 0.54$  & $40.36 \pm 2.36$ & $45.24 \pm 4.13$  & $46.86 \pm 0.60$    \\ 
        Task 4                    & $78.20 \pm 1.01$ & $77.48 \pm 1.04$  & $60.38 \pm 1.02$  & $31.06 \pm 1.38$ & $35.56 \pm 3.65$  & $33.66 \pm 0.18$ \\
        Task 5                    & $78.90 \pm 0.69$ & $76.94 \pm 0.79$  & $63.02 \pm 0.47$  & $36.84 \pm 1.60$ & $41.68 \pm 2.94$  & $36.78 \pm 0.64$ \\
         Task 6                    & $78.16 \pm 0.59$ & $76.74 \pm 0.77$  & $63.24 \pm 0.85$  & $41.50 \pm 3.36$ & $40.60 \pm 1.24$  & $41.66 \pm 0.57$ \\
          Task 7                    & $77.64 \pm 0.97$ & $76.24 \pm 0.82$  & $63.48 \pm 0.72$  & $47.64 \pm 2.14$ & $48.02 \pm 2.80$  & $33.88 \pm 0.62$ \\
           Task 8                   & $74.98 \pm 0.42$ & $76.78 \pm 0.77$  & $63.46 \pm 1.64$  & $47.68 \pm 3.54$ & $53.48 \pm 2.15$  & $40.54 \pm 1.09$ \\
            Task 9                    & $80.12 \pm 0.41$ & $77.70 \pm 1.14$  & $62.99 \pm 1.13$  & $62.26 \pm 2.54$ & $60.26 \pm 5.10$  & $58.72 \pm 0.53$ \\
             Task 10                    & $86.88 \pm 0.69$ & $88.36 \pm 0.37$  & $79.84 \pm 1.09$  & $69.62 \pm 3.27$ & $67.92 \pm 6.51$  & $63.40 \pm 0.58$ \\
        \bottomrule \\
    \end{tabular}
    \label{tab:cifar100_exact_replay}
\end{table*}

\begin{figure}[H]
\centering
\subfloat[]{\includegraphics[width=1.5in]{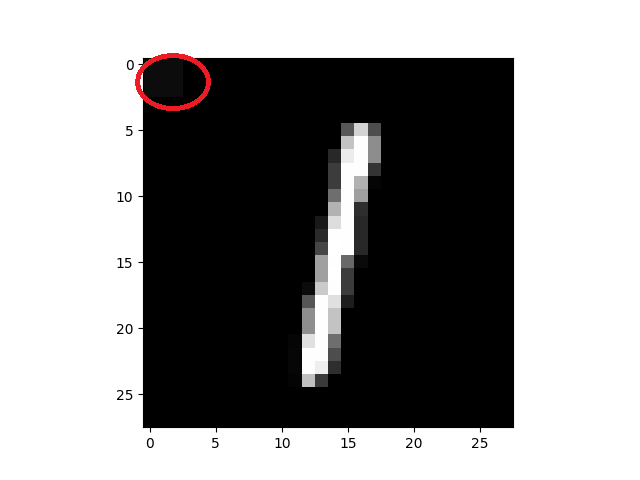}%
\label{img_att_patt_mnist}}
\hfil
\subfloat[]{\includegraphics[width=1.5in]{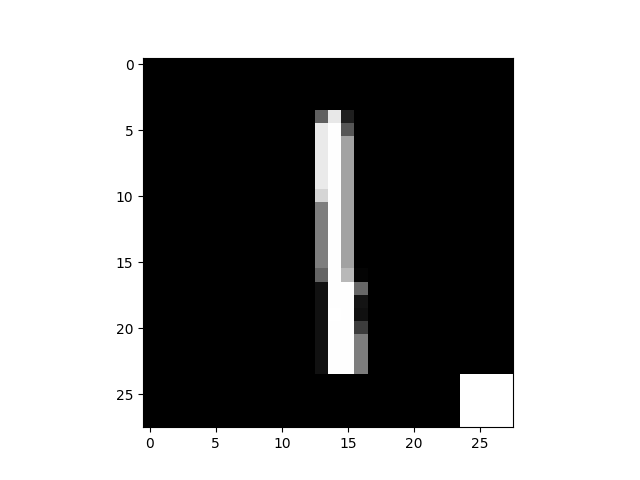}%
\label{img_def_patt_mnist}}
\hfil
\caption{Imperceptible attack pattern \& perceptible defense pattern for MNIST images: (a) image containing imperceptible square pattern as an attacker's pattern, the red circle is added later only to show the location of the attack pattern, and is not part of the actual image; (b) image containing perceptible square pattern as a defender's  pattern.}
\label{fig:mnist_inv_bd}
\end{figure}

We also consider the continual variant of MNIST dataset. Similar to CIFAR-10 dataset, MNIST also consists of 5 different tasks, where each task is a binary classification problem. Same as before, we assume that the attacker's target task is Task 1 with class 0 as the attacker's desired false label. The attacker provides a small amount of malicious samples into the training data of its target task. The malicious samples contain an imperceptible pattern, which is a square pattern inserted at the top left corner of the image. Sample image with attacker's imperceptible pattern is shown in Fig. \ref{img_att_patt_mnist}. For the MNIST dataset, the smaller imperceptible pattern is enough to achieve 100\% attack success rate, however, with our adversary aware continual learning framework, the attack performance on MNIST dataset significantly drops (significant increase in the robust performance) as demonstrated later in this section. Note that the red circle is added only to highlight the location of the attacker's imperceptible pattern as it is not possible to visually see the pattern through human eye. The red circle does not exist in the actual attack samples.

To counter this attack, the defender provides additional defensive samples in to the training data. As the defender is unaware about the target task and the target class, 
the defensive samples span all possible classes seen thus far including both current and previous task(s). For MNIST dataset, the defender provides 500 clean (correctly labeled) defensive samples per class for each task into the training data. Note that clean MNIST dataset contains more than 5000 samples per class per task. The defensive samples contain the perceptible defensive pattern, which similar to CIFAR-10 and CIFAR-100 datasets, is a white square pattern added at the bottom right corner of the image.
Sample image with defender's perceptible pattern is shown in Fig. \ref{img_def_patt_mnist}. Note that unbeknownst to the defender, the perceptible pattern for MNIST dataset does not overlap with the attacker's imperceptible pattern, which further demonstrate the promising nature of our proposed Adversary Aware Continual Learning (AACL) framework. AACL reasonably improves the robust accuracy of the model even when there is no overlap between the attacker's imperceptible pattern and the defensive perceptible pattern.

Table \ref{tab:mnist_exact_replay} shows the individual tasks' test time accuracy for both ITAML and RPS-net on MNIST dataset.  We note that the improvement in robust accuracy is more significant from the attack performance. For both ITAML and RPS-net, our proposed AACL defensive framework achieves robust (defense) accuracy closer to the clean accuracy for the target task. For ITAML, the accuracy increases from 49.73\% to 95.01\% and for RPS-net, the accuracy increases from 47.97\% to 98.05\%. Note that the attack success rate for this MNIST based continual dataset is approximately 100\% but our defensive framework completely eliminates the impact of the attack causing the attack success rate to drop to 0\%.

\begin{table*}[t]
    \centering
    \caption{
        Test accuracy (in \%) of Exact Replay-based class incremental learning approaches on MNIST. Performances on Task 1 is in bold
    }
    \begin{tabular}{lcccccc}
        \multicolumn{7}{c}{} \\ \toprule
            & \multicolumn{3}{c}{ITAML}   & \multicolumn{3}{c}{RPS-net} \\ \cmidrule(lr){2-4} \cmidrule(lr){5-7}
        Tasks &   Clean   & Attack    & AACL Defense    & Clean & Attack   & AACL Defense     \\ \cmidrule(lr) {1-1} \cmidrule(lr){2-4} \cmidrule(lr){5-7}
        Task 1                    & $\mathbf{98.99} \pm \mathbf{0.64}$ & $\mathbf{49.73} \pm \mathbf{2.10}$  & $\mathbf{95.01} \pm \mathbf{3.26}$ & $\mathbf{95.29} \pm \mathbf{0.49}$ & $\mathbf{47.97} \pm \mathbf{3.82}$  & $\mathbf{98.05} \pm \mathbf{0.21}$   \\
        Task 2                 & $95.71\pm 0.78$ & $96.94 \pm 0.57$ & $96.96 \pm 1.72$  & $74.87 \pm 0.86$ & $75.25 \pm 0.72$ & $86.44 \pm 1.25$  \\
        Task 3                    & $97.57 \pm 0.78$ & $98.75 \pm 0.33$  & $98.73 \pm 0.33$  & $74.94 \pm 0.81$ & $75.75 \pm 1.20$  & $84.13 \pm 0.95$    \\ 
        Task 4                    & $96.69 \pm 1.19$ & $97.05 \pm 1.42$  & $98.12 \pm 1.01$  & $88.49 \pm 0.66$ & $88.52 \pm 1.43$  & $92.82 \pm 1.09$ \\
        Task 5                    & $96.60 \pm 0.76$ & $98.32 \pm 0.32$  & $98.51 \pm 0.51$  & $99.21 \pm 0.11$ & $99.21 \pm 0.17$  & $98.99 \pm 0.09$ \\
        \bottomrule \\
    \end{tabular}
    \label{tab:mnist_exact_replay}
\end{table*}

\subsubsection{Defending Other Tasks}
In order to show that our proposed adversary aware continual learning (AACL) defensive framework improves the robust accuracy of the class-incremental learning algorithms regardless of which task being attacked by the attacker, we run the same AACL framework with the exact same setting different times, each time picking a different target task and target class as the attacker's desired target task and target class respectively.
Table \ref{tab:cifar10_exact_replay_OT} shows that for each target task, our proposed AACL framework reasonably improves the test time accuracy on ITAML and RPS-net using CIFAR-10 dataset.
We report the attack and defense performances for different cases in Table \ref{tab:cifar10_exact_replay_OT}, where each case representing a different target task picked by the attacker. It can be seen that regardless of which task is being attacked by the attacker, our defensive framework increases the test time performance to about 30-40\% for each target task. 

Table \ref{tab:cifar100_exact_replay_OT} shows the similar results for the more challenging CIFAR-100 dataset on both ITAML and RPS-net algorithms. It can be seen again that our proposed defensive framework immensely improves the robust accuracy of exact replay-based algorithms regardless of the target task. For most of the tasks, the improvement in performance is about 20-30\% more than the attack performance. At minimum, the increase in test time performance is about 9\% for this complex dataset.

We also run the same experiments for simpler continual variant of MNIST dataset and the results are shown in Table \ref{tab:mnist_exact_replay_OT}. It can be seen that the improvement in robust (defense) accuracy for both of the exact replay-based algorithms is considerable for all the tasks. More specifically, our proposed defensive framework provides 20\% minimum improvement in the robust (AACL defense) accuracy from the attack setting. 

We note that specifically for RPS-net algorithm when evaluated on Task 2 to Task 10 of CIFAR-100 dataset, the improvement in defense accuracy on later tasks starting from Task 8 is not as considerable as it is for the previous tasks. One possible reason is the exposure of the model to an increasingly less number of defensive samples from the later task(s) as compared to the previous task. A plausible solution to improve the accuracy on the later task is to add an additional task-dependent parameter that progressively provides more defensive samples as the model sees more and more tasks in the future.
\begin{table*}[t]
    \centering
    \caption{
        Test accuracy (in \%) of Exact Replay-based class incremental learning approaches on CIFAR-10. Attack performances and defense performances are shown for task 2 to task 5
    }
    \begin{tabular}{lcccc}
        \multicolumn{5}{c}{} \\ \toprule
            & \multicolumn{2}{c}{ITAML}   & \multicolumn{2}{c}{RPS-net} \\ \cmidrule(lr){2-3} \cmidrule(lr){4-5}
         Target Task   & Attack    & AACL Defense     & Attack   & AACL Defense     \\ \cmidrule(lr) {1-1} \cmidrule(lr){2-3} \cmidrule(lr){4-5}
        Task 2                    & $44.30$ & $\mathbf{76.89}$  & 
        $26.20$  &
        $\mathbf{40.60}$  
        \\
        Task 3                 & $50.00$ & $\mathbf{83.50}$ & $23.50$  & $\mathbf{52.61}$ \\
        Task 4                    & $49.80$ & $\mathbf{89.50}$  & $38.55$  & $\mathbf{74.90}$     \\ 
        Task 5                    & $48.95$ & $\mathbf{86.55}$  & $29.85$  & $\mathbf{52.45}$  \\
        \bottomrule \\
    \end{tabular}
    \label{tab:cifar10_exact_replay_OT}
\end{table*}

\begin{table*}[t]
    \centering
    \caption{
        Test accuracy (in \%) of Exact Replay-based class incremental learning approaches on CIFAR-100. Attack performances and defense performances are shown for task 2 to task 5
    }
    \begin{tabular}{lcccc}
        \multicolumn{5}{c}{} \\ \toprule
            & \multicolumn{2}{c}{ITAML}   & \multicolumn{2}{c}{RPS-net} \\ \cmidrule(lr){2-3} \cmidrule(lr){4-5}
         Target Task   & Attack    & AACL Defense     & Attack   & AACL Defense     \\ \cmidrule(lr) {1-1} \cmidrule(lr){2-3} \cmidrule(lr){4-5}
        Task 2                    & $12.40$ & $\mathbf{39.80}$  & 
        $1.40$  &
        $\mathbf{19.60}$  
        \\
        Task 3                 & $20.40$ & $\mathbf{48.40}$ & $3.90$  & $\mathbf{28.60}$ \\
        Task 4                    & $16.80$ & $\mathbf{46.09}$  & $1.25$  & $\mathbf{20.31}$     \\ 
        Task 5                    & $16.99$ & $\mathbf{45.00}$  & $6.73$  & $\mathbf{20.50}$  \\
        Task 6                    & $16.20$ & $\mathbf{46.50}$  & $2.50$  & $\mathbf{25.60}$  \\
        Task 7                    & $12.69$ & $\mathbf{50.80}$  & $4.90$  & $\mathbf{23.45}$  \\
        Task 8                    & $15.89$ & $\mathbf{45.60}$  & $6.10$  & $\mathbf{16.95}$  \\
        Task 9                    & $11.40$ & $\mathbf{43.19}$  & $4.30$  & $\mathbf{13.77}$  \\
        Task 10                    & $9.30$ & $\mathbf{43.90}$  & $4.11$  & $\mathbf{13.10}$  \\
        \bottomrule \\
    \end{tabular}
    \label{tab:cifar100_exact_replay_OT}
\end{table*}

\begin{table*}[t]
    \centering
    \caption{
        Test accuracy (in \%) of Exact Replay-based class incremental learning approaches on MNIST. Attack performances and defense performances are shown for task 2 to task 5
    }
    \begin{tabular}{lcccc}
        \multicolumn{5}{c}{} \\ \toprule
            & \multicolumn{2}{c}{ITAML}   & \multicolumn{2}{c}{RPS-net} \\ \cmidrule(lr){2-3} \cmidrule(lr){4-5}
         Target Task   & Attack    & AACL Defense     & Attack   & AACL Defense     \\ \cmidrule(lr) {1-1} \cmidrule(lr){2-3} \cmidrule(lr){4-5}
        Task 2                    & $57.94$ & $\mathbf{ 85.11}$  & 
        $44.02$  &
        $\mathbf{89.76}$  
        \\
        Task 3                 & $59.78$ & $\mathbf{99.15}$ & $43.73$  & $\mathbf{67.72}$ \\
        Task 4                    & $53.07$ & $\mathbf{91.13}$  & $50.85$  & $\mathbf{73.14}$     \\ 
        Task 5                    & $58.19$ & $\mathbf{78.72}$  & $53.00$  & $\mathbf{83.76}$  \\
        \bottomrule \\
    \end{tabular}
    \label{tab:mnist_exact_replay_OT}
\end{table*}

\subsection{Defending Generative Replay-Based Class Incremental Learning Approaches}

Generative replay-based continual learning approaches generate the pseudo-samples from the previous task(s) and replay these samples with the training data of the current task to achieve continual learning. These approaches are useful as they remove the necessity of storing the exact (original) samples from the previous task(s) however, their success is limited to simple MNIST based continual datasets. These approaches fail for more complex datasets primarily due to the inability of generator to generate high quality samples from the previous tasks even without any attack \cite{shen2020generative, lesort2019generative}. Therefore, the vulnerabilities of these generative replay based approaches are only considered for continual variant of MNIST dataset \cite{umer2020targeted, umer2022false} in the literature. 

We show that our proposed AACL defensive mechanism also significantly improves the robust performance of generative replay-based continual learning approaches. As before, we first assume that the attacker's target task is Task 1 with class 0 as its desired target label. The attacker inserts a small amount of malicious samples containing attacker's imperceptible pattern (a square pattern added at the top left corner of the image) in to the training data of target task. To defend this attack, the defender also provides small amount of additional defensive samples in to the training data. It is important to re-emphasize again that the defender is not aware of the target task and target class of the attacker, therefore the defensive samples consists of a set of all possible classes seen thus far. As before, we provide 500 clean (correctly labeled) defensive samples per class per task in to the training data. The defensive samples contain the defensive perceptible pattern, which is a white square pattern added at the bottom right corner of the image.

Table \ref{tab:mnist_gen_replay} shows the individual task's test time performance of two generative replay based continual learning approaches (Deep generative replay (DGR), and deep generative replay with distillation (DGR with distillation)) under three different settings, i.e., i) clean (no attack); ii) attack (with attack only); iii) AACL defense (with attack and defense) evaluated on the continual variant of MNIST dataset. We see from Table \ref{tab:mnist_gen_replay} that our proposed defensive mechanism significantly improves -- and  in fact completely recovers -- the test time performance on the attacker's target task, Task 1. More specifically, the attack success rate drops to almost 0\% with our defensive mechanism thus achieving 100\% robust (defense) performance for both of the generative replay-based class incremental learning algorithms, i.e., DGR and DGR with distillation. More specifically, the defense accuracy increases from 42.39\% to 91.06\% on DGR, and the defense accuracy is increased from 44.18\% to 94.66\% for DGR with distillation.

\subsubsection{Defending Other Tasks} We also evaluate the test time performance when other tasks are being targeted by the attacker and the results are presented in Table \ref{tab:mnist_gen_replay_OT}, which shows that our proposed defensive mechanism (AACL defense) considerably improves the test time performance for different target tasks. The improvement in robust performance from the attack scenario to defense scenario is approximately 20\% for all the cases, which shows the promising nature of our proposed adversarial aware continual learning (AACL) defensive framework. We also note that the increase in defense performance from the clean performance for the later tasks is not as considerable as it is for the previous tasks because the model sees less number of defensive samples from the later tasks. As mentioned before, one possibility to improve the defense performance on later tasks is to progressively provide more defensive samples in the later tasks via a task dependent parameter.

\begin{table*}[t]
    \centering
    \caption{
        Test accuracy (in \%) of Generative Replay-based class incremental learning approaches on MNIST. Performances on Task 1 is in bold
    }
    \begin{tabular}{lcccccc}
        \multicolumn{7}{c}{} \\ \toprule
            & \multicolumn{3}{c}{DGR}   & \multicolumn{3}{c}{DGR with distillation} \\ \cmidrule(lr){2-4} \cmidrule(lr){5-7}
        Tasks &   Clean   & Attack    & AACL Defense    & Clean & Attack   & AACL Defense     \\ \cmidrule(lr) {1-1} \cmidrule(lr){2-4} \cmidrule(lr){5-7}
        Task 1                    & $\mathbf{89.41} \pm \mathbf{1.64}$ & $\mathbf{42.39} \pm \mathbf{0.53}$  & $\mathbf{91.06} \pm \mathbf{1.89}$ & $\mathbf{95.29} \pm \mathbf{0.49}$ & $\mathbf{44.18} \pm \mathbf{0.88}$  & $\mathbf{94.66} \pm \mathbf{0.64}$   \\
        Task 2                 & $88.20\pm 0.78$ & $85.65 \pm 1.02$ & $90.68 \pm 0.30$  & $89.87 \pm 0.46$ & $86.24 \pm 0.59$ & $90.45 \pm 0.26$  \\
        Task 3                    & $88.24 \pm 2.03$ & $86.35 \pm 0.91$  & $88.08 \pm 0.85$  & $88.94 \pm 0.31$ & $89.14 \pm 0.76$  & $91.57 \pm 1.51$    \\ 
        Task 4                    & $95.17 \pm 0.19$ & $94.75 \pm 0.36$  & $96.02 \pm 0.12$  & $96.49 \pm 0.35$ & $96.15 \pm 0.18$  & $96.37 \pm 0.29$ \\
        Task 5                    & $97.18 \pm 0.32$ & $97.28 \pm 0.19$  & $96.33 \pm 0.16$  & $96.91 \pm 0.21$ & $96.94 \pm 0.28$  & $94.55 \pm 1.03$ \\
        \bottomrule \\
    \end{tabular}
    \label{tab:mnist_gen_replay}
\end{table*}

\begin{table*}[t]
    \centering
    \caption{
        Test accuracy (in \%) of Generative Replay-based class incremental learning approaches on MNIST. Attack performances and defense performances are shown for task 2 to task 5
    }
    \begin{tabular}{lcccc}
        \multicolumn{5}{c}{} \\ \toprule
            & \multicolumn{2}{c}{DGR}   & \multicolumn{2}{c}{DGR with distillation} \\ \cmidrule(lr){2-3} \cmidrule(lr){4-5}
         Target Task   & Attack    & AACL Defense     & Attack   & AACL Defense     \\ \cmidrule(lr) {1-1} \cmidrule(lr){2-3} \cmidrule(lr){4-5}
        Task 2                    & $53.09$ & $\mathbf{88.74}$  & 
        $57.30$  &
        $\mathbf{77.91}$  
        \\
        Task 3                 & $58.48$ & $\mathbf{76.41}$ & $57.47$  & $\mathbf{73.00}$ \\
        Task 4                    & $59.82$ & $\mathbf{74.97}$  & $54.25$  & $\mathbf{72.05}$     \\ 
        Task 5                    & $58.90$ & $\mathbf{71.41}$  & $53.56$  & $\mathbf{76.60}$  \\
        \bottomrule \\
    \end{tabular}
    \label{tab:mnist_gen_replay_OT}
\end{table*}

\section{Conclusion}
A novel defensive framework is proposed in this work to defend class incremental learning approaches against recent serious and insidious imperceptible backdoor attacks. We term our defensive framework as adversary aware continual learning (AACL). 
The framework similar to the imperceptible backdoor attacks, forces the class incremental learning models to learn a pattern, more specifically, defensive pattern, which is i) entirely different than the attacker's unknown imperceptible pattern and, ii) stronger (perceptible) than the attacker's pattern, using a small amount of defensive (decoy) samples. The attacker aims to associate it's imperceptible pattern to it's chosen false target label however, the defender's goal is to associate the defensive pattern to the true label. After training, when both patterns are presented to the model in the same test sample, the model pays more attention to the stronger (defensive) pattern to make it's decision and ignores the weaker (attacker's) pattern and thus correctly classifies the test sample.

We demonstrate the success of our proposed defensive framework using various class incremental learning algorithms and datasets. Moreover, we show that our proposed framework can defend the class incremental learning algorithms regardless of which task or what time step the attack happens. We believe that this is our first critical step to ensure robustness in practical continual learning algorithm, which is of paramount importance to achieve the goal of artificial general intelligence (AGI).


\bibliographystyle{IEEEtran}

\bibliography{bibliography}{}

\EOD

\end{document}